\documentclass[conference]{IEEEtran}
\IEEEoverridecommandlockouts
\usepackage{cite}
\usepackage{amsmath,amssymb,amsfonts}
\usepackage{algorithmic}
\usepackage{graphicx}
\usepackage{subfigure}
\usepackage{textcomp}
\usepackage{xcolor}
\usepackage{booktabs}
\usepackage{multirow}
\usepackage{multicol}
\usepackage[ruled,linesnumbered]{algorithm2e}
\usepackage{makecell}
\usepackage{bbding}
\usepackage{colortbl}
\usepackage{mathrsfs}  
\usepackage{pifont}  

\usepackage{tabularx}

\usepackage{amsthm}     
\usepackage{tcolorbox}
\tcbuselibrary{theorems}

\newtheorem{definition}{Definition}[subsection]

\tcolorboxenvironment{theorem}{
  boxrule=0.5pt,
  boxsep=0pt,
  left=5pt,
  right=5pt,
  top=5pt,
  bottom=5pt,
  colback=white 
}

\tcolorboxenvironment{definition}{
  boxrule=0.5pt,
  boxsep=0pt,
  left=5pt,
  right=5pt,
  top=5pt,
  bottom=5pt,
  colback=white 
}

\def\BibTeX{{\rm B\kern-.05em{\sc i\kern-.025em b}\kern-.08em
    T\kern-.1667em\lower.7ex\hbox{E}\kern-.125emX}}
\begin{document}

\title{GCA-BULF: A Bottom-Up Framework for Short-Term Load Forecasting Using Grouped Critical Appliances}


\author{Yunhao Yao, Jinwei Fang, Puhan Luo, Zhiqiang Wang, Jiahui Hou and Xiang-Yang Li,~\IEEEmembership{Fellow,~IEEE} \\
University of Science and Technology of China (USTC)
}

\maketitle

\begin{abstract}
With the rise of time-of-use and tiered electricity pricing, energy consumers are encouraged to adopt peak-shifting strategies by automatically controlling high-power appliances.
These help lower energy costs while enhancing the power grid's stability.
To support such energy management with high resilience and responsiveness, reliable short-term load forecasting (STLF) plays a critical role.
STLF predicts electricity consumption over time horizons ranging from minutes to days, using historical data, temporal patterns, and contextual factors.
Traditional top-down forecasting methods struggle to capture the complex consumption patterns of diverse and mixed appliance loads.
Although bottom-up methods improve forecasting accuracy by integrating appliance-level data, monitoring all appliances is costly, and many do not meaningfully impact total load prediction.
Therefore, we propose GCA-BULF, a bottom-up short-term load forecasting framework based on grouped critical appliances, supported by three key designs.
First, the Critical Appliance Filtering module ranks appliances according to their power consumption, switching frequency, and usage pattern periodicity, and identifies critical ones through iterative load decomposition.
Next, the Related Appliance Grouping module clusters these appliances based on spatial and temporal correlations for group-level forecasting.
Finally, the Collaborative Load Forecasting module refines the total load prediction by combining multiple group-level forecasts.
We evaluate GCA-BULF on residential and office building load forecasting tasks.
Experimental results reveal that GCA-BULF improves hourly total load forecasting by $20.85\%$-$57.88\%$ compared to existing top-down methods and by $33.03\%$-$92.48\%$ compared to bottom-up methods.
\end{abstract}


\section{Introduction}
Time-of-Use Pricing divides a day into peak, off-peak and valley periods, assigning different electricity rates.
It encourages users to shift consumption to off-peak or valley times (e.g. nighttime) for balancing grid supply and demand~\cite{yang2015electric}.
Tiered electricity pricing charges higher rates at higher usage levels, promoting energy conservation and more efficient power use~\cite{zhang2018robustly}.
With advances in Internet of Things (IoT) and edge computing, real-time monitoring and intelligent control of electrical appliances have become feasible~\cite{yao2023traffic,yao2024secoinfer,yao2025trafficdiary}.
This enables cost reduction under the above pricing schemes while improving grid stability~\cite{chakraborty2022cost}.
In this context, reliable load forecasting plays a critical role in ensuring the quality of service (QoS) in energy management systems, enabling strategies like peak-shifting, load-shedding, and storage optimization~\cite{qin2024deciding, wang2024improving}.

\textbf{Existing top-down forecasting falls short for mixed appliance loads.} 
Traditional top-down load forecasting methods predict future energy consumption by capturing short-term~\cite{rafi2021short,ijaz2022novel,lin2022short,sekhar2023robust,asiri2024short,jalalifar2024sac} or long-short-term~\cite{kiprijanovska2020houseec,yang2022combined,jiang2021hybrid,xiao2023short} temporal patterns in the total load of energy consumers (e.g. residences and office buildings).
However, due to the lack of a fine-grained understanding of load composition, the aggregate load, formed by numerous mixed appliances, exhibits complex variation patterns, making it challenging to identify consistent trends.

\textbf{Existing bottom-up forecasting overlooks appliance usage correlations.} 
Subsequent bottom-up methods enhance total load prediction by aggregating appliance-level loads to better capture total load composition~\cite{razghandi2020residential,zhou2022appliance,razghandi2021short,liu2023home,gao2018bottom,zheng2019kalman,wang2021bottom}.
However, due to appliance additions, removals and relocations, continuous monitoring of all appliances is impractical, leading to incomplete total load representation. 
Although some approaches incorporate historical total loads to fill this gap, they still overlook inter-appliance correlations~\cite{langevin2023efficient}.

Therefore, there is a pressing need for an STLF framework which can accurately estimate total energy consumption based on partial appliance data and historical total load information.
To explore this framework, we face the following challenges: \\
\textbf{(1) How to select partial critical appliances.}
Although the widespread use of smart meters makes appliance-level load monitoring feasible, tracking every appliance is still costly and impractical. 
To address this, it is essential to continuously monitor a targeted subset of appliances that strongly shape total load trends, balancing deployment complexity with forecasting accuracy. \\
\textbf{(2) How to assess inter-appliance correlations.}
Although appliance-level loads typically exhibit simpler variation patterns than total loads, they may lose important contextual information about user behaviours. 
Since user behaviours often involve the coordinated use of multiple appliances (e.g., making lunch might include: turning on kitchen lights; using a rice cooker and microwave afterwards), it is necessary to capture these usage associations to enhance appliance-level load prediction and further improve total load forecasting. \\
\textbf{(3) How to integrate appliance and total loads.}
Since the combined loads of all monitored appliances still fall short of representing the total load, simply summing appliance-level forecasts is insufficient for accurate total load prediction.
Therefore, a dedicated framework is needed to refine the total load forecast by integrating appliance-level predictions.
Since forecasting each appliance’s load separately lacks contextual richness, it is also essential to explore how to incorporate appliance correlations into the framework.

To address the above challenges, we propose GCA-BULF.
To our knowledge, it is the first bottom-up short-term load forecasting framework that selects critical appliances and incorporates appliance correlations to enhance total load prediction.
The key contributions are summarized as follows:

\noindent~\textbullet~A Critical Appliance Filtering module that selects a minimal set of appliances carrying sufficient information to capture total load trends.
Based on power consumption, state change frequency, and usage pattern periodicity, we quantify each appliance's contribution to total load variation. 
By iteratively decomposing the loads of high-contribution appliances, we smooth the residual loads and identify a critical appliance subset preserving the key trends in total load variation.

\noindent~\textbullet~A Related Appliance Grouping module that organizes critical appliances according to their usage correlations.
We quantify spatial and temporal relationships using lagged correlation, yielding a unified usage-correlation measure for each appliance pair.
By clustering with this measure as the distance metric, we group critical appliances with strong associations.

\noindent~\textbullet~A Collaborative Load Forecasting framework that enhances total load prediction through two-stage inference. 
In stage one, each critical appliance group is forecasted independently, and a preliminary total load estimation is produced using temporal features enhanced by discrete wavelet transform (DWT).
In stage two, the forecasts from multiple appliance groups are integrated to refine the final total load prediction.

\noindent~\textbullet~GCA-BULF achieves strong performance on both residential and office building load forecasting tasks.
We implement a prototype of GCA-BULF and evaluate it on two datasets: the widely used residential energy consumption dataset UK-DALE~\cite{kelly2015uk}, and our self-collected office building dataset, BST-EC.
Experimental results show that GCA-BULF reduces forecasting errors by $20.85\%$–$57.88\%$ compared to existing top-down methods, and by $33.03\%$–$92.48\%$ compared to existing bottom-up methods.

\section{Related Works}
In this section, we classify existing short-term load forecasting methods into two categories: top-down and bottom-up, and provide a comprehensive review of each.

\subsection{Top-Down Short-Term Load Forecasting}
Top-down forecasting relies on historical loads from energy consumers (e.g., residences, office buildings) and incorporates various influencing factors (e.g., weather, weekday, time of day) to predict future loads at the minute or hour level~\cite{eren2024comprehensive, ullah2024short, ma2023review, akhtar2023short, rodrigues2023short}.
According to the span of historical loads utilized, top-down methods can be further divided into short-term data-driven and long-short-term integrated methods.

\textbf{Short-Term Data-Driven} methods assume that energy consumption patterns remain relatively consistent over short time horizons. 
Therefore, they employ deep neural networks such as CNN-LSTM~\cite{rafi2021short,jalalifar2024sac}, FC-LSTM~\cite{ijaz2022novel}, and CNN-BiLSTM~\cite{sekhar2023robust,asiri2024short} to extract spatial features and temporal patterns from recent historical data. 
Some studies further incorporate hyperparameter optimization~\cite{sekhar2023robust,asiri2024short} and attention mechanisms~\cite{lin2022short} to improve performance.

\textbf{Long-Short-Term Integrated} methods assume that energy consumption exhibits both short-term consistency and long-term cyclicality.
Approaches such as HousEEC~\cite{kiprijanovska2020houseec}, Co-LSTM~\cite{yang2022combined}, and MFDL~\cite{jiang2021hybrid} combine recent short-term historical loads with long-term loads from the same time periods in previous weeks for forecasting.
Noteably, Xiao et al.~\cite{xiao2023short} first construct a long-term energy consumption profile and then use it to refine predictions derived from short-term features.

\subsection{Bottom-Up Short-Term Load Forecasting}
Since appliance loads typically follow simpler consumption patterns, bottom-up forecasting first predicts the future loads of individual appliances and then aggregates them to achieve a more accurate total load forecast~\cite{eren2024comprehensive, ullah2024short, ma2023review, akhtar2023short, rodrigues2023short}.
Some existing studies focus on improving the accuracy of appliance-level load forecasting, while others explore aggregating historical loads of multiple appliances to enhance total load prediction.

\textbf{Appliance-Level Load Forecasting} focuses on predicting the future loads of individual appliances. The total load forecast can then be obtained by suming the predicted appliance-level loads.
Razghandi et al.~\cite{razghandi2020residential, razghandi2021short} and Liu et al.~\cite{liu2023home} formulate the forecasting problem as a sequence-to-sequence task and employ encoder–decoder architectures to address it, while Zhou et al. design an error estimation network to refine predictions produced by the LSTM-FC model~\cite{zhou2022appliance}.

\textbf{Appliance-Based Collaborative Load Forecasting} learns temporal patterns from historical appliance loads and integrates them during inference to improve the accuracy of total load prediction.
Early studies primarily relied on statistical approaches.
Gao et al.~\cite{gao2018bottom} and Zheng et al.~\cite{zheng2019kalman} estimate each appliance's state with statistical models and then aggregate the expected total load.
In contrast, Wang et al.~\cite{wang2021bottom} and Langevin et al.~\cite{langevin2023efficient} apply deep learning methods to capture nonlinear variations in appliance loads, achieving a more accurate alignment with the overall load curve.

\subsection{Limitations}
Top-down short-term load forecasting relies solely on total historical load data, which simplifies data collection. However, the complexity of consumption patterns makes accurate forecasting more challenging.
Bottom-up methods provide deeper insights into energy consumption by utilizing detailed appliance‑level data.
However, continuously monitoring all appliances in real‑world settings is impractical, as appliances may be added, removed, or relocated.
Moreover, not all appliances have a significant impact on total load forecasting; low-power or consistently operating appliances often contribute little to prediction accuracy.

\begin{figure*}[t]
    \centering
    \includegraphics[width=1.7\columnwidth]{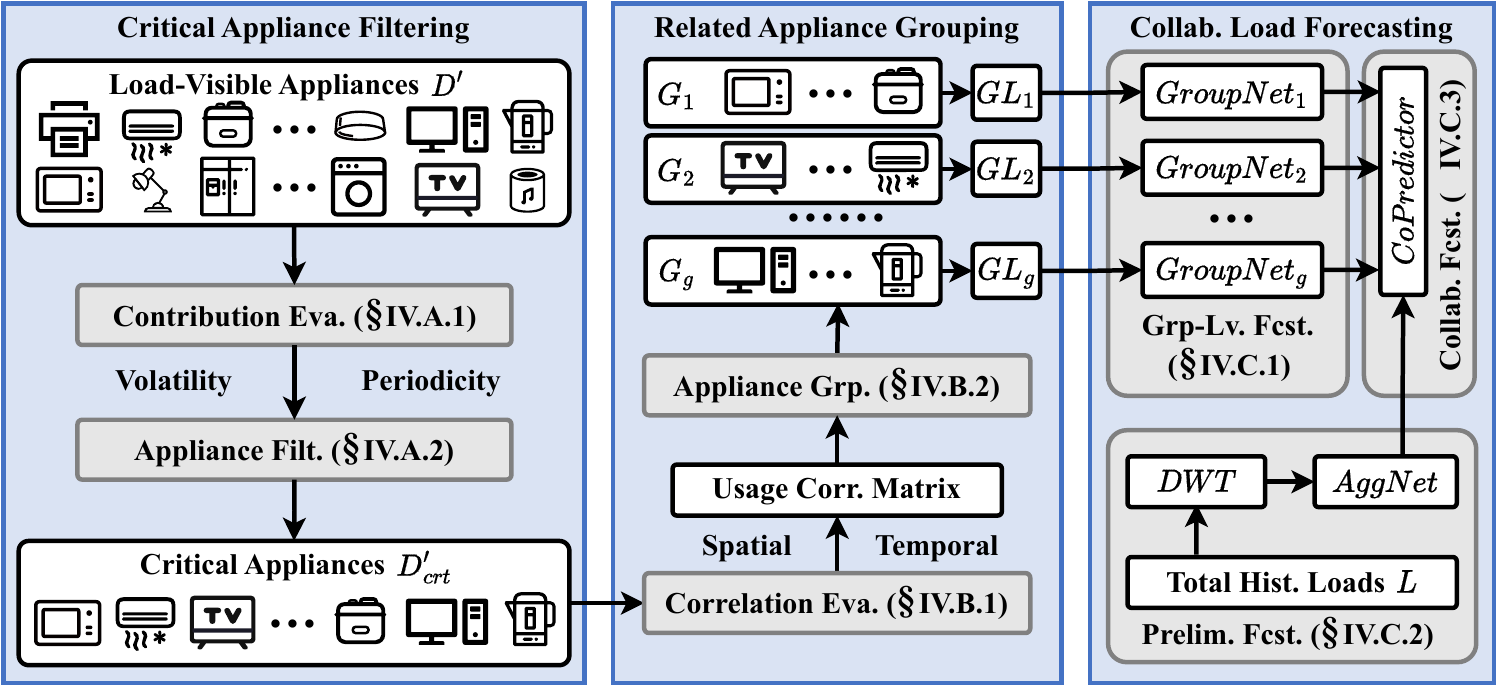}
    \caption{System Overview of GCA-BULF.}
    \label{fig:system}
\end{figure*}

\section{Problem Statement}
We consider a typical energy consumption scenario, such as a residence or an office building, where all appliances form a set $D$. Since some appliances may be added, removed or relocated, real-time monitoring of all appliances' loads is costly and impractical.
Therefore, we define a subset $D'=\{d_1, d_2, ..., d_n\} \subseteq D$, representing the appliances whose loads can be monitored.
Let $L=[l_1, l_2, ..., l_m]$ denote the total load at each time step for all appliances in $D$, and $L_i=[l^i_1, l^i_2, ..., l^i_m] (1 \leq i \leq n)$ represent the load of appliance $d_i$.
Then, for any time step $t \in [1, m]$, the total monitored load satisfies: $\sum_{i=1}^n l^i_t \leq l_t$.
Our task is to predict future energy consumption, leveraging recent loads from monitorable appliances and integrating them with recent total loads.
The task is formally defined as follows:
$$
\mathcal{F} = \arg\min_{F} ||l_{m+1} - F(L[-\tau:],L_1[-\tau:], ..., L_n[-\tau:])||_2,
$$
where $\tau$ represents the length of the time window.

\section{Methodology}

In this section, we provide a detailed explanation of GCA-BULF.
First, the \textit{Critical Appliance Filtering} module identifies critical appliances by evaluating the volatility and periodicity of their load patterns.
According to spatial and temporal usage correlations, the \textit{Related Appliance Grouping} module clusters these critical appliances into groups.
Finally, by integrating multiple group-level loads along with the total load, GCA-BULF performs collaborative short-term load forecasting for smart spaces, such as individual residences or entire buildings.
The system overview of GCA-BULF is presented in Fig.~\ref{fig:system}.

\subsection{Critical Appliance Filtering}

\subsubsection{Contribution Evaluating}
We note that the monitored appliance loads in $D'$ affect total load trends differently, which can be viewed from three perspectives:

\noindent~\textbullet~\textbf{Power Consumption}: High-power appliances (e.g., air conditioners) influence total load fluctuations much more than low-power ones (e.g., lights), as their usage introduces more pronounced changes in overall energy consumption.

\noindent~\textbullet~\textbf{State Change Frequency}: Appliances that frequently switch states (e.g., kettles) introduce more corresponding fluctuations to the total loads compared to those with stable operation (e.g., refrigerators). Capturing this correspondence can better assist in total load forecasting.

\noindent~\textbullet~\textbf{Usage Pattern Periodicity}: Appliances with more periodic usage patterns (e.g., rice cookers) have more predictable loads than those with irregular usage (e.g., televisions). 
Assisting in total load forecasting with highly periodic appliances can reduce misleading information.

Since the range of load fluctuations reflects an appliance's power consumption, and the frequency of fluctuations indicates its state changes, we use the normalized variance of historical loads to model these fluctuations as follows:
$$
\bar{l^i} = \frac{1}{m}\sum_{j=1}^m l^i_j, \, \, var(L_i) = \frac{1}{m}\sum_{j=1}^m (l^i_j - \bar{l^i})^2,
$$
$$
vola(d_i) = Norm(var(L_i)) \in [0, 1],
$$
where $Norm(\cdot)$ denotes a min-max normalization function.
The measure $vola(\cdot)$ reflects both the power consumption level and the frequency of state changes of each appliance.

The within-group sum of squares (WGSS) quantifies the total variation of samples from their group mean, indicating the degree of discreteness within the group.
To mitigate the impact of energy consumption magnitude differences on usage periodicity evaluation, we normalize the historical load sequence $L_i$ for each appliance $d_i$.
Assuming that each day contains $k$ data points, the normalized load sequence $Norm(L_i)=[Norm(l^i_1), ..., Norm(l^i_m)] \in [0, 1]^m$ can then be reorganized into daily load segments as $DL_i = [dl^i_1, ..., dl^i_\frac{m}{k}] \in [0, 1]^{k \times \frac{m}{k}}$.
By treating $DL_i$ as a sample group, we use the WGSS to quantify the discreteness in appliance $d_i$'s daily power consumption patterns.
Therefore, the usage periodicity of $d_i$ can be characterized using the normalized WGSS as follows:
$$
\bar{dl^i} = \frac{k}{m}\sum_{j=1}^\frac{m}{k} dl^i_j, \, \, WGSS(DL_i) = \sum_{j=1}^\frac{m}{k} ||dl^i_j - \bar{dl^i}||^2,
$$
$$
period(d_i) = 1 - Norm(WGSS(DL_i)) \in [0, 1],
$$
where $Norm(\cdot)$ denotes a min-max normalization function.
A larger value of $period(\cdot)$ indicates more regular usage.
By combining $vola(\cdot)$ and $period(\cdot)$, we define each appliance’s contribution $ctrb(\cdot)$ to total load forecasting in \textbf{Definition~\ref{def:app_im}}.
Since $vola(\cdot)$ captures both power consumption and state change frequency, we recommend a trade-off factor $\alpha < 1$ to place greater emphasis on $vola(\cdot)$. 

\begin{definition}[Appliance Contribution]
\label{def:app_im}
    For appliance $d_i$, higher power consumption or more frequent state changes result in a larger $vola(d_i)$, while more regular usage patterns lead to a higher $period(d_i)$. Therefore, we define $ctrb(d_i) = vola(d_i) + \alpha \cdot period(d_i)$ to represent the contribution of $d_i$'s loads $L_i$ on the total load forecasting.
\end{definition}

\begin{figure}[t]
    \centering
    \includegraphics[width=0.9\columnwidth]{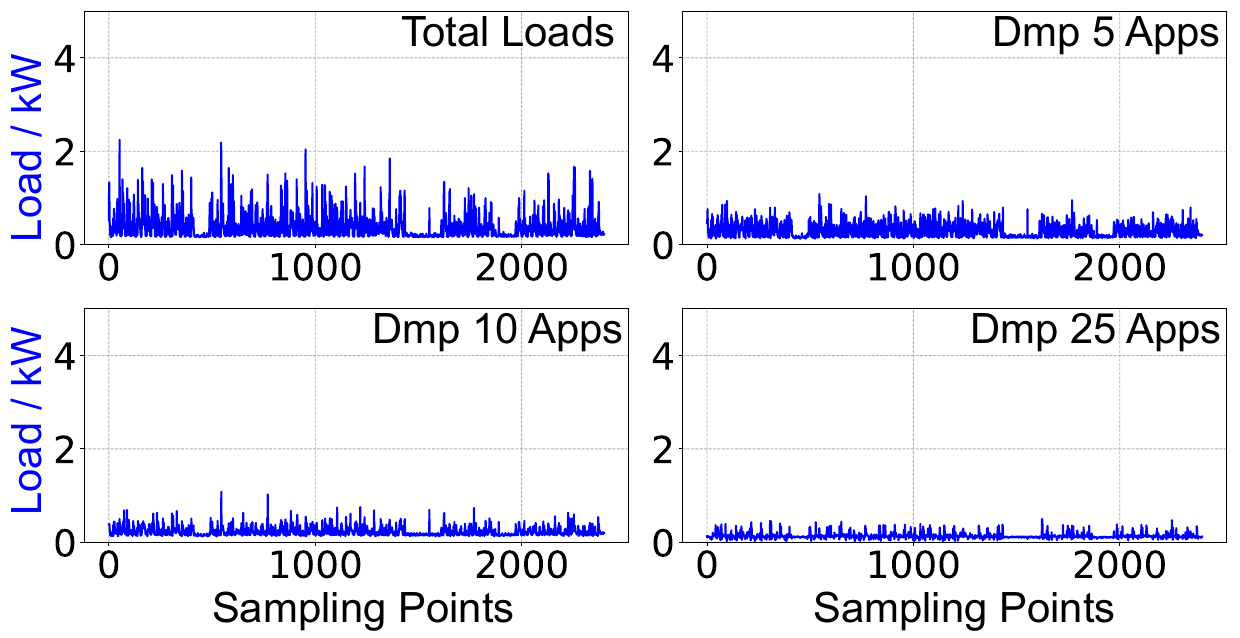}
    \caption{Examples of Critical Appliance Filtering on the UK-DALE Dataset.}
    \label{fig:app_dmp_examples}
\end{figure}

\subsubsection{Appliance Filtering}
After sorting the appliances in $D'$ by their contribution $ctrb(\cdot)$ in descending order, we iteratively decompose the load sequence of each high-contribution (critical) appliance from the total load sequence $L$ until the standard deviation of the remaining loads falls below a threshold $\sigma$ and finally derive a set of critical appliances, denoted as $D'_{crt}=\{d_{{crt}_1}, ..., d_{{crt}_\eta}\} \subseteq D'$.
This approach ensures that once a sufficient number of appliance-level loads have been decomposed, fluctuations in the residual loads become negligible. 
This indicates that the appliance loads in $D'_{crt}$ capture enough information to represent the trend of total load variations.
Therefore, critical appliances in $D'_{crt}$ constitute the minimum feasible deployment. Compared with monitoring all appliances, this can significantly reduce deployment overhead.
Fig.~\ref{fig:app_dmp_examples} presents an example from the UK-DALE dataset, illustrating how the residual load curve evolves as the loads of critical appliances are progressively filtered from $L$.

\subsection{Related Appliance Grouping}

\begin{figure}[t]
    \centering
    \includegraphics[width=0.95\columnwidth]{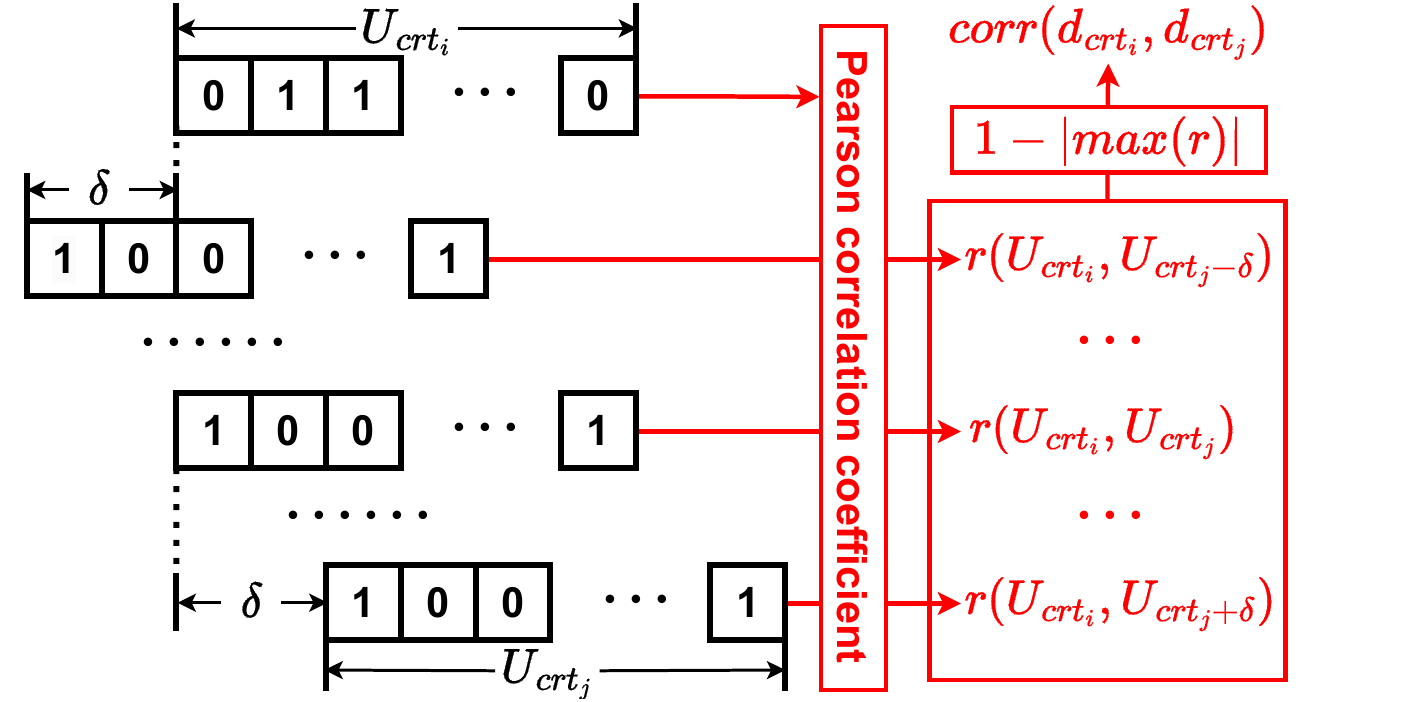}
    \caption{Examples of Calculating the Correlation of Critical Appliance Usage.}
    \label{fig:app_usage_corr_examples}
\end{figure}

\begin{figure*}[t]
    \centering
    \includegraphics[width=1.7\columnwidth]{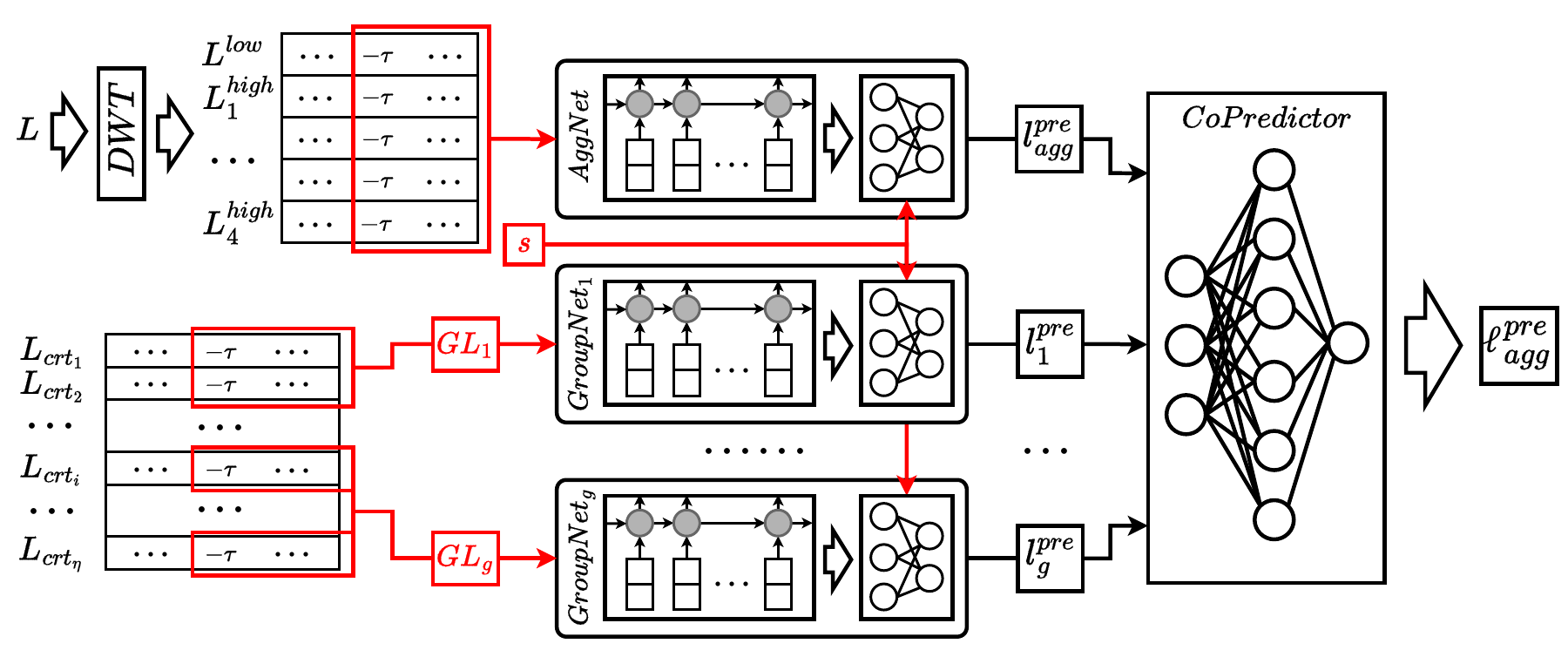}
    \caption{The Structure of Our Collaborative Load Forecasting Framework.}
    \label{fig:coll_forecast_net}
\end{figure*}

\subsubsection{Correlation Evaluating}
Bottom-up load forecasting relies on more predictable appliance-level loads, refining total load prediction. 
Therefore, improving the accuracy of appliance-level forecasting benefits total load forecasting performance.
Specifically, we group critical appliances based on their usage correlations. This serves two purposes: reducing overfitting and prediction complexity by replacing appliance-level forecasting with group-level forecasting; enhancing group-level forecasting accuracy by incorporating appliance usage correlations.
Appliance usage is typically correlated in two main aspects:

\noindent~\textbullet~\textbf{Spatial (linkage) correlation}: Appliances located in the same area may exhibit related usage patterns. 
For example, microwaves and rice cookers are commonly used together during meal preparation. Such correlations arise from specific user behaviours, leading to simultaneous appliance usage.

\noindent~\textbullet~\textbf{Temporal (causal) correlation}: User behaviours often follow a sequential pattern, creating causal relationships between appliance usage.
For instance, a user entering the bedroom before sleeping may first turn on the ceiling light and TV, then switch to a bedside lamp before going to bed. This sequence establishes a temporal dependency in appliance operation.

To capture these relationships, we employ the lagged correlation to quantify the dependency between each pair of critical appliances.
The lagged correlation measures the relationship between two time series by assessing their correlation at different time offsets.
By varying the lag, we can align two time series and determine the maximum correlation coefficient between them. 
In this context, a zero time lag characterizes the linkage relationship, indicating simultaneous usage, while a positive or negative time lag reflects the causal relationship, capturing sequential dependencies in appliance usage.
We give the calculation of the lag correlation as follows:
$$
C_{\delta}(\mathbf{x}, \mathbf{y}) = \max\limits_{\Delta t \in [-\delta, \delta]} \frac{\sum_t (\mathbf{x}_t - \bar{\mathbf{x}})(\mathbf{y}_{t+\Delta t} - \bar{\mathbf{y}})}{\sqrt{\sum_t (\mathbf{x}_t - \bar{\mathbf{x}})^2 \cdot \sum_t (\mathbf{y}_{t+\Delta t} - \bar{\mathbf{y}})^2}},
$$
where $\mathbf{x}$ and $\mathbf{y}$ represent two time series, and $\delta$ denotes the time lag.
Following~\cite{zheng2019kalman}, we extract the usage vector $U_{{crt}_k}$ to represent state changes for each critical appliance $d_{{crt}_k} \in D'_{crt}$, which is defined as: 
$$
U_{{crt}_k} = [\mathbb{I}(l^{{crt}_k}_i > \frac{\max(L_{{crt}_k})}{20})]_{i=1}^m,
$$ 
where $\frac{\max(L_{{crt}_k})}{20}$ is the threshold used to separate the idle and active states of $d_{{crt}_k}$, based on its historical loads $L_{{crt}_k}$ as defined in \cite{zheng2019kalman}.
Combined with the lag correlation, we define the usage correlation between each pair of critical appliances in \textbf{Definition~\ref{def:app_corr}}.
Fig.~\ref{fig:app_usage_corr_examples} provides an example of how the correlation of appliance usage is calculated.

\begin{definition}[Appliance Usage Correlation]
\label{def:app_corr}
    For critical appliances $d_{crt_i}, d_{crt_j} \in D'_{crt}$, $U_{crt_i}$ and $U_{crt_j}$ denote their respective usage vectors. 
    We define the correlation between their usage patterns by the lag correlation as:
    $$corr(d_{crt_i}, d_{crt_j}) = 1 - |C_\delta(U_{crt_i}, U_{crt_j})| \in [0, 1],$$
    where a lower $corr(d_{crt_i}, d_{crt_j})$ indicates a stronger correlation, serving as the distance function in Section  IV.B.2.
\end{definition}

\subsubsection{Appliance Grouping}
We group critical appliances in $D'_{crt}$ based on their usage correlation to enhance load forecasting, optimizing it from the appliance-level to the group-level. 
The key insight is that grouping strongly correlated appliances allows prediction models to capture both spatial and temporal dependencies, leading to more accurate group-level load forecasting.
Specifically, we set the time lag $\delta$ to 6 hours, dividing a day into four periods: morning, afternoon, evening, and early morning. 
This choice allows $corr(\cdot, \cdot)$ to capture temporal correlations in appliance usage within the same period.
Subsequently, we employ the DBSCAN algorithm to cluster the critical appliances as follows:
$$
\begin{aligned}
\{G_1, ..., G_g\} = DBSCAN(D'_{crt}, \epsilon, MinPts=1), \\
\end{aligned}
$$
where $corr(\cdot, \cdot)$ defines the distance function, $\epsilon$ is the experimentally determined neighborhood radius, and $MinPts=1$ ensures that no critical appliance is excluded as noise.
Moreover, $D'_{crt} = G_1 \cup ...\cup G_g$ with $G_i \cap G_j = \emptyset$ for $i \neq j$.

\subsection{Collaborative Load Forecasting}

\subsubsection{Group-Level Load Forecasting}
For each group $G_i (1 \leq i \leq g)$ of critical appliances, we define a matrix $GL_i \in \mathbb{R}^{\tau \times \eta_i}$ recording their historical loads over a time window of length $\tau$, where $\eta_i$ is the number of appliances in $G_i$. 
For each group $G_i$, we construct an LSTM-FC structured network, denoted as $GroupNet_i(\cdot)$, to perform group-level forecasting.
The LSTM component of $GroupNet_i(\cdot)$ is computed as follows:
$$ 
F_t = ReLU(W_f[H_{t-1}, (GL_i)_{t}] + b_f), 
$$
$$
I_t = ReLU(W_i[H_{t-1}, (GL_i)_{t}] + b_i), 
$$
$$ 
\widetilde{C}_t = tanh(W_c[H_{t-1}, (GL_i)_{t}] + b_c),
$$
$$
C_t = F_t \odot C_{t-1} + I_t \odot \widetilde{C}_t, 
$$
$$ 
O_t = ReLU(W_o[H_{t-1}, (GL_i)_{t}] + b_o), 
$$
$$
H_t = O_t \odot tanh(C_t), 
$$
where, at the $t$-th time step, $F_t, I_t, \widetilde{C}_t, C_t, O_t,$ and $H_t$ denote the forget gate, input gate, candidate memory cell, memory cell, output gate, and hidden state, respectively.
The subsequent fully connected (FC) component generates the group-level forecasting results as follows:
$$ 
h^{(0)}_i = [O_\tau, s], 
$$
$$
h^{(\iota)}_i = ReLU(W^{(\iota)} h^{(\iota-1)}_i + b^{(\iota)}), 
$$
$$
l^{pre}_i = h^{(n_l)},
$$
where $s$ denotes additional influencing factors and $n_l$ is the number of hidden layers. In our experiments, $s$ includes weekday and time‑of‑day indicators.

\begin{table*}[t]
    \centering
    \caption{MAE (KW) and MAPE of GCA-BULF with Varying Clustering Radius $\epsilon$ and Time Windows $\tau$ on UK-DALE}
    \begin{tabularx}{\textwidth}{
    >{\centering\arraybackslash}p{0.052\textwidth} |
    >{\centering\arraybackslash}p{0.06\textwidth} |
    >{\centering\arraybackslash}p{0.06\textwidth} |
    >{\centering\arraybackslash}p{0.06\textwidth} |
    >{\centering\arraybackslash}p{0.06\textwidth} |
    >{\centering\arraybackslash}p{0.06\textwidth} |
    >{\centering\arraybackslash}p{0.06\textwidth} |
    >{\centering\arraybackslash}p{0.06\textwidth} |
    >{\centering\arraybackslash}p{0.06\textwidth} |
    >{\centering\arraybackslash}p{0.06\textwidth} |
    >{\centering\arraybackslash}p{0.06\textwidth} |
    >{\centering\arraybackslash}p{0.06\textwidth}
    }
    \toprule
    \midrule
    \large{$\epsilon$} & 0.3 & 0.4 & 0.5 & 0.6 & 0.7 & 0.8 & 0.9 & 0.92 & 0.94 & 0.96 & 0.98 \\
    \midrule
    \multirow{2}{*}{$\tau = 3$} & 0.0927 & 0.0927 & 0.0912 & 0.0905 & 0.0879 & 0.0874 & 0.0853 & \textbf{0.0834} & 0.0841 & 0.0882 & 0.0906 \\
    \cmidrule{2-12}
    & 18.61\% & 18.69\% & 19.07\% & 18.77\% & 17.77\% & 17.94\% & 17.40\% & \textbf{17.02\%} & 17.11\% & 18.29\% & 18.77\% \\
    \midrule
    \multirow{2}{*}{$\tau = 6$} & 0.0890 & 0.0872 & 0.0859 & 0.0851 & 0.0827 & 0.0807 & 0.0785 & \textbf{0.0761} & \textbf{0.0761} & 0.0800 & 0.0815 \\
    \cmidrule{2-12}
    & 18.06\% & 17.54\% & 18.03\% & 17.15\% & 16.94\% & 16.92\% & 16.21\% & \textbf{15.76\%} & 15.84\% & 16.58\% & 16.82\% \\
    \midrule
    \multirow{2}{*}{$\tau = 12$} & 0.0822 & 0.0817 & 0.0796 & 0.0764 & 0.0741 & 0.0734 & 0.0733 & \textbf{0.0700} & 0.0702 & 0.0749 & 0.0794 \\
    \cmidrule{2-12}
    & 16.86\% & 17.17\% & 16.37\% & 15.86\% & 15.54\% & 15.42\% & 15.49\% & \textbf{14.50\%} & \textbf{14.50\%} & 15.80\% & 16.50\% \\
    \midrule
    \bottomrule
    \end{tabularx}
    \label{tab:time_step_th_ukdale}
\end{table*}

\subsubsection{Preliminary Total Load Forecasting}
We apply the discrete wavelet transform (DWT) to decompose the total historical loads $L$ into four high-frequency components $(L^{high}_1, L^{high}_2, L^{high}_3, L^{high}_4)$ and one low-frequency component $(L^{low})$. 
This decomposition captures the multi‑scale characteristics of complex load patterns in energy consumption and reduces noise.
Similarly, the components are segmented with a $\tau$-length time window, and an LSTM–FC‑based prediction network $AggNet(\cdot)$ is constructed for preliminary total load forecasting:
$$
L = L^{low} \oplus L^{high}_1 \oplus L^{high}_2 \oplus L^{high}_3 \oplus L^{high}_4,
$$
$$
L^{DWT} = [L^{low}; L^{high}_1; L^{high}_2; L^{high}_3; L^{high}_4], 
$$
$$
l^{pre}_{agg} = FC([LSTM(L^{DWT}[-\tau:])_\tau, s]).
$$
Since the LSTM and FC components of $AggNet(\cdot)$ and $GroupNet_i(\cdot)$ share the same structure, we omit the details of $AggNet(\cdot)$ for brevity.

\subsubsection{Collaborative Total Load Forecasting}
We develop our collaborative forecasting framework by synergizing $g$ specialized $GroupNet(\cdot)$ instances with an $AggNet(\cdot)$, as shown in Fig.~\ref{fig:coll_forecast_net}. 
$AggNet(\cdot)$ provides a preliminary total load forecast, while $GroupNet_1(\cdot), ..., GroupNet_g(\cdot)$ predict the future loads of all critical appliance groups significantly affecting total load fluctuations.
By integrating the group‑level forecasts with the preliminary total load prediction, $CoPredictor(\cdot)$ refines the total load estimation as follows:
$$ 
h^{(0)}_i = [l^{pre}_{agg}, l^{pre}_1, l^{pre}_2, ..., l^{pre}_{g}],
$$
$$
h^{(\iota)}_i = ReLU(W^{(\iota)} h^{(\iota-1)}_i + b^{(\iota)}), 
$$
$$
\ell^{pre}_{agg} = h^{(n_l)}. 
$$
The framework follows a two-stage training process: independent pre-training of $AggNet(\cdot)$ and each $GroupNet_i(\cdot)$, followed by joint fine-tuning of $CoPredictor(\cdot)$ with the pre-trained components.

\section{Experiments}

We evaluate GCA-BULF on a residence-level and a building-level load dataset, comparing its performance against state-of-the-art top-down and bottom-up forecasting methods.
The key highlights of our evaluation are as follows:

\textbullet~The three core modules-\textit{Critical Appliance Filtering}, \textit{Related Appliance Grouping}, and \textit{Collaborative Load Forecasting}-along with the DWT-based feature enhancement, significantly improve the performance of GCA‑BULF (Section V.B).

\textbullet~GCA-BULF achieves lower hourly forecasting errors than existing methods.
Specifically, it improves forecast performance by $20.85\%$–$57.88\%$ over top-down methods and by $33.03\%$–$92.48\%$ over bottom-up methods (Section V.C).

\subsection{Experiment Setup}
The UK Domestic Appliance-Level Electricity (UK-DALE)~\cite{kelly2015uk} dataset, collected by Imperial College, is a widely used benchmark for non-intrusive load monitoring and energy consumption analysis.
It contains power consumption data from five UK residences, recorded across 2012 to 2015 at a sampling rate of 1/6 Hz. 
The dataset includes both aggregate and appliance-level power readings for over 50 household appliances, such as refrigerators, washing machines, dishwashers, and microwave ovens, enabling detailed load disaggregation and energy usage analysis.

Additionally, we collected building energy consumption data from 2022 to 2024 at the Basic Science Teaching Experimental Centre of a Chinese university, which we refer to as the BST-EC dataset.
The BST-EC dataset includes total energy consumption, along with lighting and office plug loads for each of the 18 floors.
These measurements are recorded hourly by a central smart meter and individual sub-meters installed in the distribution room.
We conduct ablation and comparative experiments on GCA-BULF using the above two datasets.

To enable evaluation of hourly short‑term load forecasting for residential and building scenarios, we preprocess the two datasets.
The UK‑DALE dataset records instantaneous power consumption every six seconds using high‑frequency smart meters.
Hourly average loads for each appliance and residence are computed by averaging these high‑frequency samples.
In contrast, the BST‑EC dataset provides hourly load data derived from differences between consecutive smart meter readings, requiring no additional preprocessing.

\begin{table*}[t]
    \centering
    \caption{MAE (10MW) and MAPE of GCA-BULF with Varying Clustering Radius $\epsilon$ and Time Windows $\tau$ on BST-EC}
    \begin{tabularx}{\textwidth}{
    >{\centering\arraybackslash}p{0.065\textwidth} |
    >{\centering\arraybackslash}p{0.067\textwidth} |
    >{\centering\arraybackslash}p{0.067\textwidth} |
    >{\centering\arraybackslash}p{0.067\textwidth} |
    >{\centering\arraybackslash}p{0.067\textwidth} |
    >{\centering\arraybackslash}p{0.067\textwidth} |
    >{\centering\arraybackslash}p{0.067\textwidth} |
    >{\centering\arraybackslash}p{0.067\textwidth} |
    >{\centering\arraybackslash}p{0.067\textwidth} |
    >{\centering\arraybackslash}p{0.067\textwidth} |
    >{\centering\arraybackslash}p{0.067\textwidth} 
    }
    \toprule
    \midrule
    \large{$\epsilon$} & 0.1 & 0.2 & 0.3 & 0.35 & 0.4 & 0.45 & 0.5 & 0.6 & 0.7 & 0.9  \\
    \midrule
    \multirow{2}{*}{$\tau = 3$} & \textbf{0.0888} & 0.0982 & 0.1053 & 0.1176 & 0.1105 & 0.1208 & 0.1135 & 0.1064 & 0.1088 & 0.1133 \\
    \cmidrule{2-11}
    & \textbf{2.18\%} & 2.39\% & 2.56\% & 2.80\% & 2.69\% & 2.93\% & 2.74\% & 2.58\% & 2.61\% & 2.72\% \\
    \midrule
    \multirow{2}{*}{$\tau = 6$} & \textbf{0.0790} & 0.0907 & 0.1097 & 0.1089 & 0.1056 & 0.1192 & 0.1139 & 0.1085 & 0.1122 & 0.1136 \\
    \cmidrule{2-11}
    & \textbf{1.94\%} & 2.25\% & 2.70\% & 2.66\% & 2.55\% & 2.83\% & 2.79\% & 2.64\% & 2.68\% & 2.75\% \\
    \midrule
    \multirow{2}{*}{$\tau = 12$} & \textbf{0.0717} & 0.0903 & 0.1027 & 0.1116 & 0.0989 & 0.1124 & 0.1023 & 0.1115 & 0.1026 & 0.1058 \\
    \cmidrule{2-11}
    & \textbf{1.79\%} & 2.21\% & 2.60\% & 2.79\% & 2.39\% & 2.78\% & 2.48\% & 2.71\% & 2.52\% & 2.57\% \\
    \midrule
    \bottomrule
    \end{tabularx}
    \label{tab:time_step_th_bstec}
\end{table*}

\begin{table*}[t]
  \centering
  \begin{minipage}[t]{0.48\textwidth}
    \centering
    \caption{MAE (KW) and MAPE of GCA-BULF with and without Critical Appliance Filtering on UK-DALE}
    \resizebox{1.0\linewidth}{!}{
    \begin{tabular}{c|c c|c c|c c}
        \toprule
        \midrule
        & \multicolumn{2}{c|}{$\tau = 3$} & \multicolumn{2}{c|}{$\tau = 6$} & \multicolumn{2}{c}{$\tau = 12$} \\
        \midrule
        Filter? & \ding{51} & \ding{55} & \ding{51} & \ding{55} & \ding{51} & \ding{55} \\
        \midrule
        MAE & 0.0834 & 0.0908 & 0.0761 & 0.0806 & 0.0700 & 0.0745  \\
        \midrule
        MAPE & 17.02\% & 18.04\% & 15.76\% & 16.31\% & 14.50\% & 15.16\% \\
        \midrule
        \bottomrule
    \end{tabular}
    }
    \label{tab:filtering_ukdale}
  \end{minipage}
  \hfill
  \begin{minipage}[t]{0.48\textwidth}
    \centering
    \caption{MAE (10MW) and MAPE of GCA-BULF with and without Critical Appliance Filtering on BST-EC}
     \resizebox{0.95\linewidth}{!}{
        \begin{tabular}{c|c c|c c|c c}
        \toprule
        \midrule
        & \multicolumn{2}{c|}{$\tau = 3$} & \multicolumn{2}{c|}{$\tau = 6$} & \multicolumn{2}{c}{$\tau = 12$} \\
        \midrule
        Filter? & \ding{51} & \ding{55} & \ding{51} & \ding{55} & \ding{51} & \ding{55} \\
        \midrule
        MAE & 0.0888 & 0.0928 & 0.0790 & 0.0862 & 0.0717 & 0.0763  \\
        \midrule
        MAPE & 2.18\% & 2.26\% & 1.94\% & 2.11\% & 1.79\% & 1.89\% \\
        \midrule
        \bottomrule
        \end{tabular}
    }
    \label{tab:filtering_bstec}
  \end{minipage}
\end{table*}

\subsection{Ablation Experiments}

\subsubsection{Evaluating Different Appliance Grouping Radius}

We adjust the clustering radius and time window in the DBSCAN-based \textit{Related Appliance Grouping} module to determine the optimal usage correlation threshold and evaluate the necessity of grouping critical appliances.
We use Mean Absolute Error (MAE) and Mean Absolute Percentage Error (MAPE) to evaluate forecasting performance.
MAE measures the absolute difference between predicted and actual values, in the same unit as the target (e.g., KW).
MAPE expresses this error as a percentage of the actual values, enabling scale-free comparison.
Experimental results on the UK‑DALE and BST‑EC datasets are presented in TABLE~\ref{tab:time_step_th_ukdale} and TABLE~\ref{tab:time_step_th_bstec}.

The results in TABLE~\ref{tab:time_step_th_ukdale} indicate that a time window of $\tau=12$ is the most suitable choice, suggesting that residential energy consumption exhibits strong short-term temporal correlations over a half-day period. 
Additionally, GCA-BULF achieves the best forecasting performance when the clustering radius is set to $\epsilon=0.92$, regardless of the selected time window. 
It significantly outperforms both $\epsilon=0.3$ and $\epsilon=0.98$, reducing MAPE by $1.59\%$–$2.36\%$ and $1.06\%$–$2.00\%$, equivalent to performance improvements of $8.54\%$-$14.00\%$ and $6.30\%$-$12.12\%$.
These results confirm the effectiveness of the \textit{Critical Appliance Grouping} module.
Specifically, $\epsilon = 0.3$ leads to each appliance being grouped individually, while $\epsilon = 0.98$ treats all appliances as a single group, disabling grouping.

Similarly, as shown in Table~\ref{tab:time_step_th_bstec}, a window of $\tau=12$ also proves to be a suitable choice for office building load forecasting. 
Notably, GCA-BULF achieves the best forecasting performance when the cluster radius is set to $\epsilon=0.1$, indicating that grouping appliances individually yields the most accurate results.
This is because the appliance-level data in BST-EC already represents group-level loads, such as all lighting or plug loads on a specific floor.
Therefore, the \textit{Critical Appliance Grouping} module does not perform additional grouping.

\begin{figure*}[t]
    \centering
    \begin{minipage}[b]{0.48\linewidth}
        \centering
        \includegraphics[width=1.0\columnwidth]{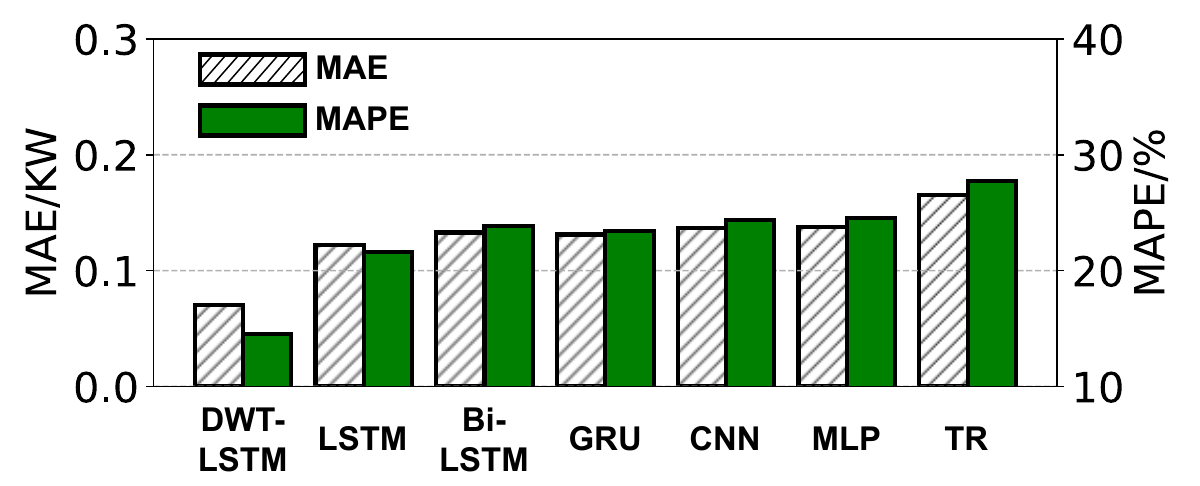}
        \caption{MAE (KW) and MAPE of GCA-BULF Under Varying Model Structures on UK-DLAE.}
        \label{fig:diff_models_ukdale}
    \end{minipage}
    \hfill
    \begin{minipage}[b]{0.48\linewidth}
        \centering
        \includegraphics[width=1.0\columnwidth]{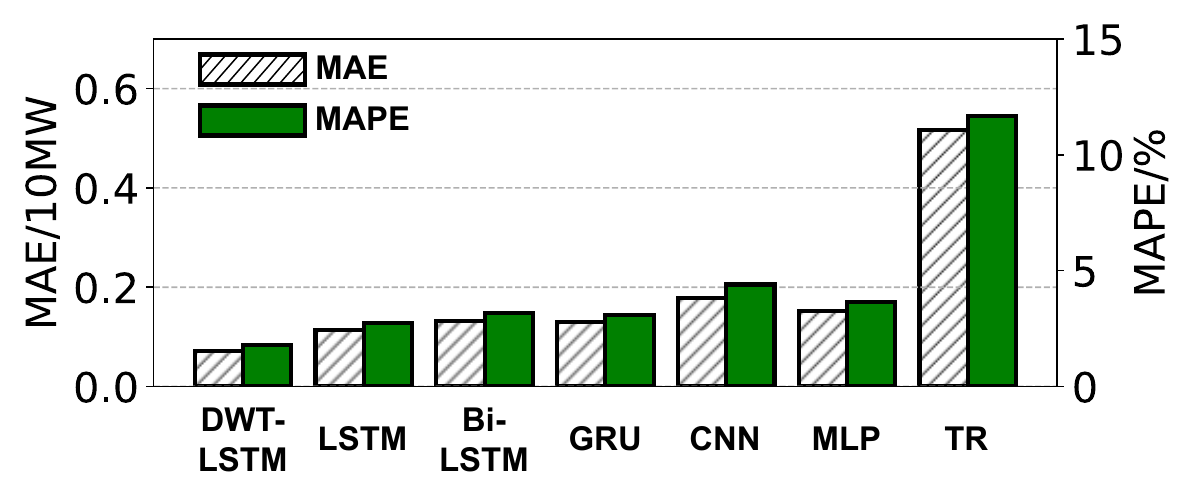}
        \caption{MAE (10MW) and MAPE of GCA-BULF Under Varying Model Structures on BST-EC.}
        \label{fig:diff_models_bstec}
    \end{minipage}
\end{figure*}

\subsubsection{Evaluating the Importance of Critical Appliance Filtering}

We evaluate the necessity of the \textit{Critical Appliance Filtering} module under various time window settings.
TABLE~\ref{tab:filtering_ukdale} and TABLE~\ref{tab:filtering_bstec} present the hourly load forecasting performance of GCA-BULF on the UK‑DALE and BST‑EC datasets, comparing cases with and without the \textit{Critical Appliance Filtering} module.

On UK-DALE, GAC-BULF with critical appliance filtering consistently lowers load forecasting errors across all time window settings. MAPE drops by $0.55\%$-$1.02\%$, corresponding to a performance improvement of $3.37\%$-$5.65\%$. 
On BST-EC, applying critical appliance filtering reduces MAPE by $0.08\%$-$0.17\%$, equivalent to a $3.54\%$-$8.06\%$ improvement.
These results clearly highlight the necessity of the \textit{Critical Appliance Filtering} module, as it effectively selects appliances with higher volatility and stronger periodicity to participate in total load forecasting.
In contrast, many low-contribution appliances exhibit near-flat loads, offering little to total load variations. Others, with weak periodicity, tend to introduce larger errors that degrade total load forecasting accuracy.

\begin{figure*}[t]
    \centering
    \begin{minipage}[b]{0.48\linewidth}
        \centering
        \includegraphics[width=1.0\columnwidth]{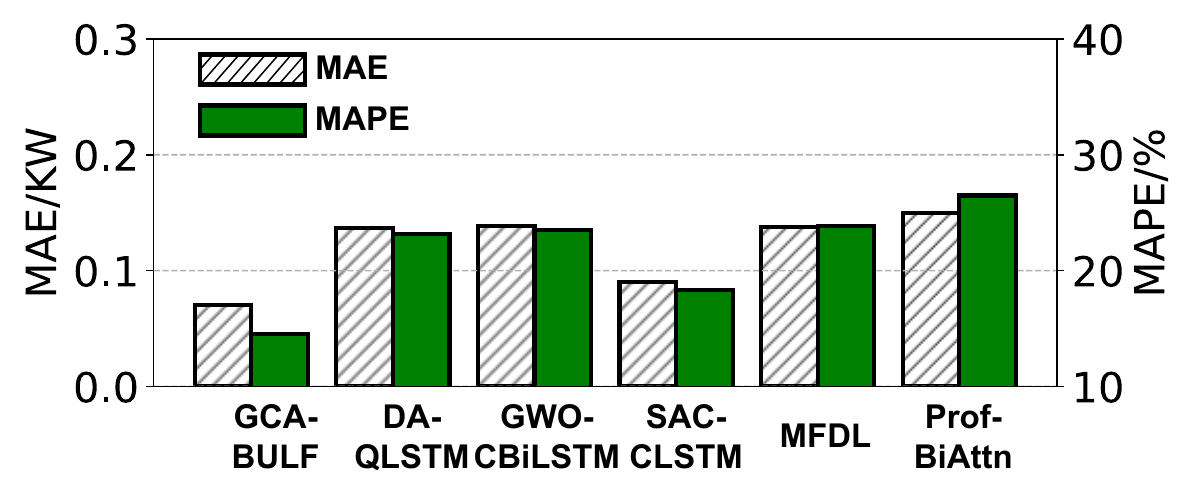}
        \caption{Comparison with Existing Top-Down Methods on UK-DALE.}
        \label{fig:top_down_ukdale}
    \end{minipage}
    \hfill
    \begin{minipage}[b]{0.48\linewidth}
        \centering
        \includegraphics[width=1.0\columnwidth]{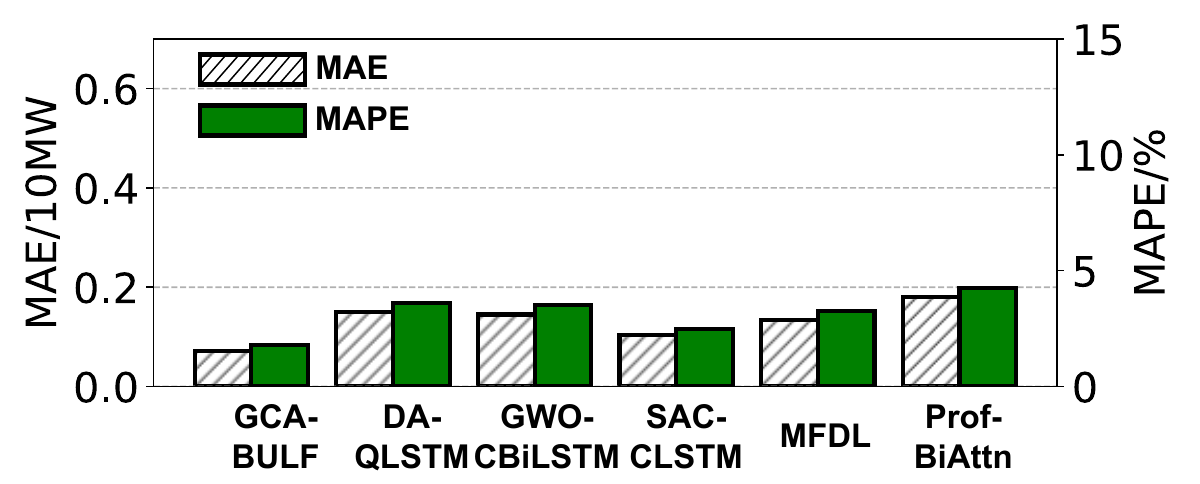}
        \caption{Comparison with Existing Top-Down Methods on BST-EC.}
        \label{fig:top_down_bstec}
    \end{minipage}
\end{figure*}

\begin{figure*}[t]
    \centering
    \includegraphics[width=2.0\columnwidth]{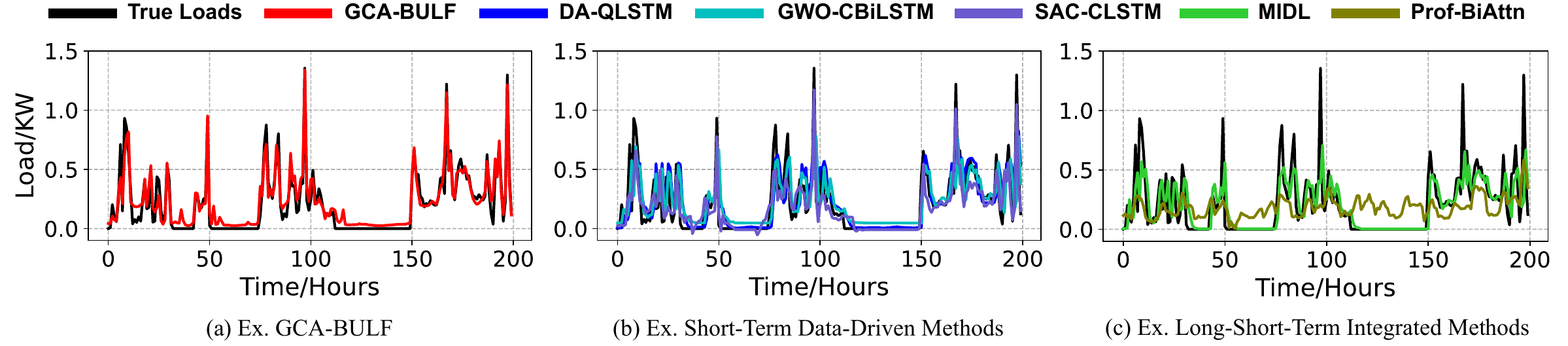}
    \caption{Case Study: GCA-BULF vs. Existing Top-Down Methods on UK-DALE.}
    \label{fig:demo_topdown_ukdale}
\end{figure*}

\subsubsection{Evaluating Different Collaborative Forecasting Models}

We modify the model architecture employed in the \textit{Collaborative Load Forecasting} module.
Fig.~\ref{fig:diff_models_ukdale} and Fig.~\ref{fig:diff_models_bstec} present the hourly load forecasting performance of GCA-BULF using different model architectures on UK-DALE and BST-EC.
We compare our proposed DWT-LSTM model with six commonly used deep learning architectures: LSTM, BiLSTM, GRU, CNN, MLP, and Transformer (TR).

GCA-BULF adopts a DWT-LSTM for preliminary total load forecasting, collaborating with group-level forecasting to refine the results.
The first two comparisons (DWT-LSTM vs. LSTM) in Figs.~\ref{fig:diff_models_ukdale} and~\ref{fig:diff_models_bstec} demonstrate that incorporating DWT enhances the representation of historical total load patterns, enabling GCA-BULF to better capture load variation trends.
Specifically, compared to LSTM alone, adding DWT reduces MAPE by $7.09\%$ on UK-DALE and $0.96\%$ on BST-EC, equivalent to $32.84\%$ and $34.90\%$ performance gains.

Additionally, by replacing LSTM with five alternative model architectures, we validate the effectiveness of selecting LSTM as the base model in the \textit{Collaborative Load Forecasting} module. 
This choice reduces MAPE by $1.86\%$–$6.15\%$ on UK-DALE and $0.33\%$–$8.92\%$ on BST-EC, corresponding to performance improvements of $7.93\%$–$22.17\%$ and $10.71\%$–$76.44\%$, respectively.
Notably, Transformers perform the worst for short-term load forecasting, likely because they favour long-term temporal dependencies and are less sensitive to rapid, short-term fluctuations common in load data.

\begin{figure*}[t]
    \centering
    \begin{minipage}[b]{0.48\linewidth}
        \centering
        \includegraphics[width=1.0\columnwidth]{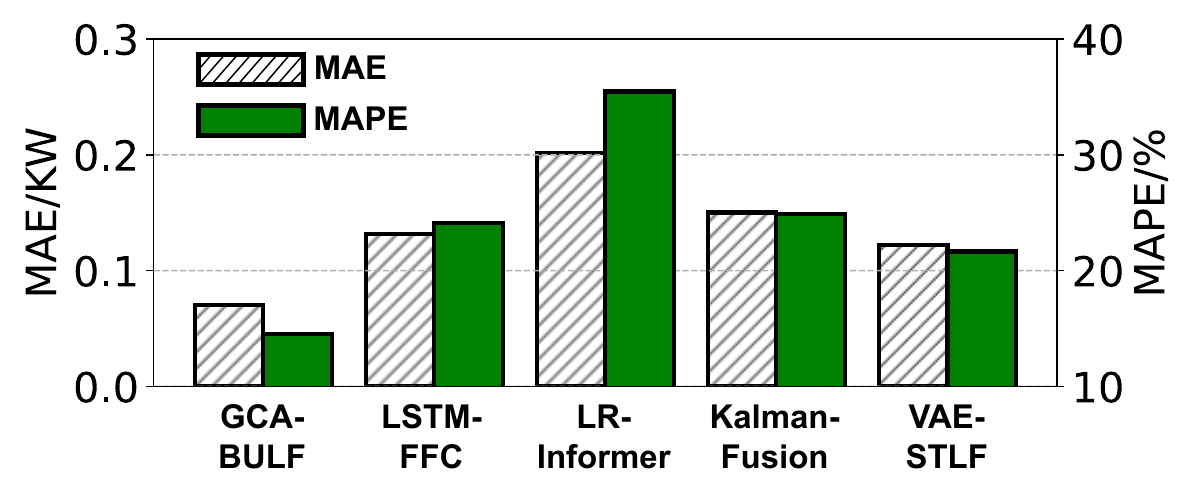}
        \caption{Comparison with Existing Bottom-Up Methods on UK-DALE.}
        \label{fig:bottom_up_ukdale}
    \end{minipage}
    \hfill
    \begin{minipage}[b]{0.48\linewidth}
        \centering
        \includegraphics[width=1.0\columnwidth]{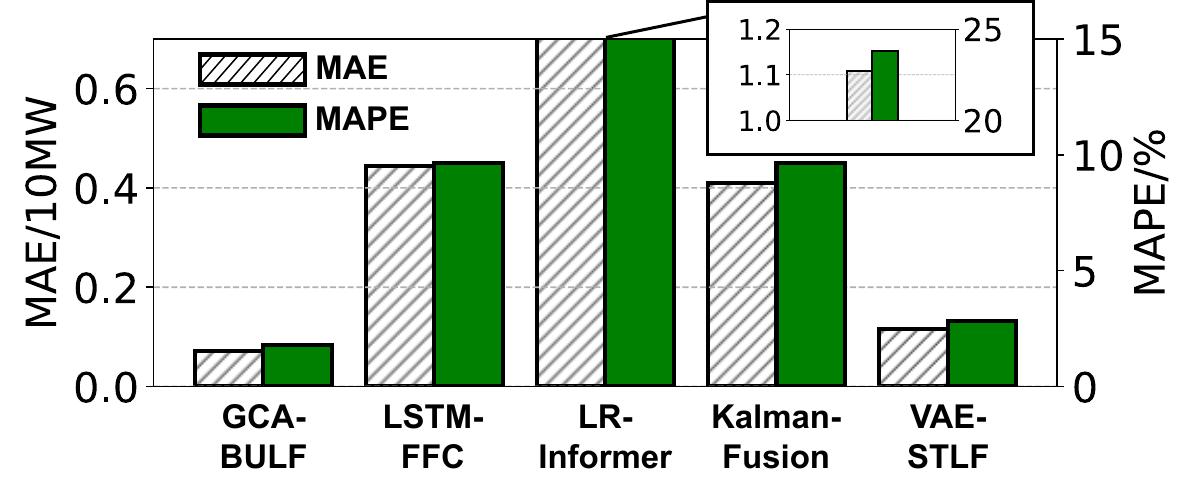}
        \caption{Comparison with Existing Bottom-Up Methods on BST-EC.}
        \label{fig:bottom_up_bstec}
    \end{minipage}
\end{figure*}

\begin{figure*}[t]
    \centering
    \includegraphics[width=2.0\columnwidth]{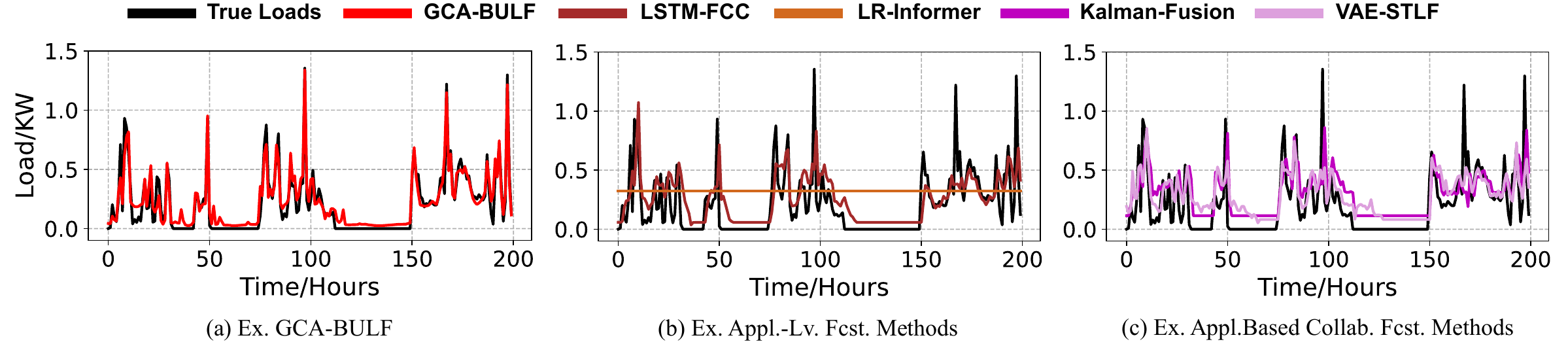}
    \caption{Case Study: GCA-BULF vs. Existing Bottom-Up Methods on UK-DALE.}
    \label{fig:demo_bottomup_ukdale}
\end{figure*}

\subsection{Comparative Experiments}

\subsubsection{Comparison with Existing Top-Down Forecasting Methods}

Top-down load forecasting methods can be further classified into two categories: short-term data-driven methods and long-short-term integrated methods. 
For the former, we compare GCA-BULF with DA-QLSTM~\cite{lin2022short} incorporating temporal attention, DWT-enhanced SAC-ConvLSTM (SAC-CLSTM)~\cite{jalalifar2024sac}, and GWO-CNN-BiLSTM (GWO-CBiLSTM)~\cite{sekhar2023robust} optimized via bio‑inspired algorithms. 
For the latter, we include MFDL~\cite{jiang2021hybrid} and Prof-BiAttn~\cite{xiao2023short}, which apply different integration mechanisms, as baseline methods.

Overall, the experimental results in Fig.~\ref{fig:top_down_ukdale} and Fig.~\ref{fig:top_down_bstec} demonstrate that GCA-BULF significantly outperforms existing top-down forecasting methods across UK-DALE and BST-EC.
On UK-DALE, GCA-BULF reduces MAPE by $3.82\%$-$11.99\%$, corresponding to a $20.85\%$-$45.26\%$ forecasting performance improvement.
On BST-EC, GCA-BULF lowers MAPE by $0.71\%$-$2.46\%$, which translates to a $28.40\%$-$57.88\%$ improvement.
As shown in Fig.~\ref{fig:demo_topdown_ukdale}, GCA-BULF’s forecast tracks the true load more closely than existing top-down methods.
Although SAC-ConvLSTM also performs well, it significantly underperforms GCA-BULF in capturing peak values.

Notably, long-short-term integrated methods underperform short-term data-driven ones, likely because residential and building loads have strong short-term correlations but weak long-term dependencies.
Among short-term methods, SAC-ConvLSTM performs best, aided by DWT-enhanced load representations.
DA-QLSTM, despite using the optimal LSTM, overweights external factors (e.g., weekdays, weather) and underuses history loads. 
GWO-CNN-BiLSTM exhibits poor performance, as its BiLSTM architecture is not an ideal choice for short-term load forecasting.
In addition to prior strengths of existing approaches, such as adopting the optimal LSTM structure and DWT-enhanced load representations, GCA-BULF further improves forecasting accuracy by incorporating critical appliance loads to refine total load prediction.

\subsubsection{Comparison with Existing Bottom-Up Forecasting Methods}

Bottom-up load forecasting methods fall into two categories: appliance-level forecasting and appliance-based collaborative forecasting.
For appliance-level methods, we select LR-Informer~\cite{liu2023home} with an encoder-decoder structure and LSTM-FCC~\cite{zhou2022appliance} with error estimation in our comparison.
For collaborative methods, we select statistics-based Kalman-Fusion~\cite{zheng2019kalman} and two-stage, appliance-to-household VAE-STLF~\cite{langevin2023efficient} as baseline methods.

Overall, Fig.~\ref{fig:bottom_up_ukdale} and Fig.~\ref{fig:bottom_up_bstec} demonstrate that GCA-BULF consistently outperforms existing bottom-up methods on both UK-DALE and BST-EC.
Specifically, on UK-DALE, GCA-BULF reduces MAPE by $7.15\%$-$20.97\%$, corresponding to a $33.03\%$-$59.12\%$ performance improvement.
On BST-EC, GCA-BULF achieves a MAPE reduction of $1.03\%$-$22.00\%$, translating to a $36.52\%$-$92.48\%$ enhancement.
As shown in Fig.~\ref{fig:demo_bottomup_ukdale}, GCA-BULF’s forecast matches true loads far better than existing bottom-up methods.
Notably, LR-Informer produces an almost flat prediction curve, again underscoring the poor applicability of the Transformer architecture for short-term load forecasting.

Both categories of approaches-predicting each appliance individually (e.g., LSTM-FCC, LR-Informer) or forecasting combined appliance loads (e.g., Kalman-Fusion)-share a fundamental limitation: appliance-level data cannot fully cover total loads. 
We formalize this limitation as a constraint in Section III.
Due to the lack of load information from unmonitored appliances, most bottom-up methods exhibit poorer performance than top-down approaches.
LR-Informer performs significantly worse, primarily because the Transformer-like Informer architecture struggles to capture short-term load dependency. 
In contrast, VAE-STLF outperforms other bottom-up methods, as it incorporates the historical total loads.
However, VAE-STLF ignores inter-appliance usage correlations. GCA-BULF captures these correlations via the \textit{Related Appliance Grouping} module, thereby improving appliance-level prediction and enhancing total load forecasting.
Besides, the \textit{Critical Appliance Filtering} module identifies appliances that contribute most to total loads, and the integration of DWT further improves feature representations from historical loads.

\section{Discussion}
With the widespread adoption of time-of-use and tiered electricity pricing, shifting the usage of high-power appliances and avoiding excessive consumption can help reduce energy costs.
In smart home and smart office scenarios equipped with diverse sensors, including fine-grained smart meters, GCA-BULF enables accurate hourly load forecasting.
This capability supports both manual and automated appliance control, facilitating resilient and responsive energy management. 

Unlike existing bottom‑up methods that monitor all appliance loads, GCA-BULF only selects a small set of critical appliances, reducing deployment costs and enabling broader applicability.
Moreover, by incorporating appliance usage correlations, GCA-BULF gains a deeper understanding of energy consumption patterns, which enhances collaborative forecasting and reduces load prediction errors.
However, GCA-BULF is customized for energy users. 
As users adjust their appliance usage in response to its forecasting guidance, the model’s structure and parameters must also adapt accordingly. 
To address this dynamic environment, future work will focus on applying online learning techniques for adaptive optimization of GCA‑BULF.

\section{Conclusion}
In this paper, we propose GCA-BULF, a bottom-up short-term load forecasting framework built upon grouped critical appliances.
To our knowledge, GCA-BULF is the first to both filter critical appliances and model inter-appliance correlations to enhance total load forecasting.
Specifically, GCA-BULF first identifies critical appliances by their contribution, reducing deployment requirements.
It then clusters appliances by usage correlations to improve appliance-level predictions.
Finally, group-level forecasting combined with DWT-enhanced load representations reduces total load forecasting error.
On real-world datasets, GCA-BULF achieves relative errors of $14.50\%$ for residential and $1.79\%$ for office building load forecasting, improving over existing top-down methods by $20.85\%$–$57.88\%$ and bottom-up methods by $33.03\%$–$92.48\%$.
In the future, we plan to extend GCA-BULF to an online, adaptive framework that continuously updates in response to changing user energy usage patterns, ensuring greater adaptability to dynamic consumption scenarios.



\bibliographystyle{IEEEtran}
\bibliography{reference.bib}

\end{document}